\documentclass[10pt,twocolumn,letterpaper]{article}

\usepackage{cvpr}
\usepackage{times}
\usepackage{epsfig}
\usepackage{graphicx}
\usepackage{amsmath}
\usepackage{amssymb}
\usepackage{romannum}

\usepackage[breaklinks=true,bookmarks=false]{hyperref}

\cvprfinalcopy 

\def\cvprPaperID{1857} 

\ifcvprfinal\fi
\begin{document}
	
	\title{A-Lamp: Adaptive Layout-Aware Multi-Patch Deep Convolutional Neural Network for Photo Aesthetic Assessment}

	\author{Shuang Ma{$^\star$},
		Jing Liu{$^\dagger$} and
		Chang Wen Chen{$^\star$}\\
		{$^\star$}Computer Science and Engineering, SUNY at Buffalo\\ {$^\dagger$}Electrical and Information Engineering, Tianjin University\\
		{\tt\small \{shuangma,chencw\}@buffalo.edu} {\tt\small,} {\tt\small jliu$\_$tju@tju.edu.cn}
	}
	\maketitle
	\begin{abstract}			
		Deep convolutional neural networks (CNN) have recently been shown to generate promising results for aesthetics assessment. However, the performance of these deep CNN methods is often compromised by the constraint that the neural network only takes the fixed-size input. To accommodate this requirement, input images need to be transformed via cropping, warping, or padding, which often alter image composition, reduce image resolution, or cause image distortion. Thus the aesthetics of the original images is impaired because of potential loss of fine grained details and holistic image layout. However, such fine grained details and holistic image layout is critical for evaluating an image's aesthetics. 
		In this paper, we present an Adaptive Layout-Aware Multi-Patch Convolutional Neural Network (A-Lamp CNN) architecture for photo aesthetic assessment. This novel scheme is able to accept arbitrary sized images, and learn from both fined grained details and holistic image layout simultaneously. To enable training on these hybrid inputs, we extend the method by developing a dedicated double-subnet neural network structure, i.e. a Multi-Patch subnet and a Layout-Aware subnet. We further construct an aggregation layer to effectively combine the hybrid features from these two subnets. Extensive experiments on the large-scale aesthetics assessment benchmark (AVA) demonstrate significant performance improvement over the state-of-the-art in photo aesthetic assessment.
		
	\end{abstract}
	
	\section{Introduction} \label{intro}
	
	\begin{figure}
		\centering
		\includegraphics[scale=0.24]{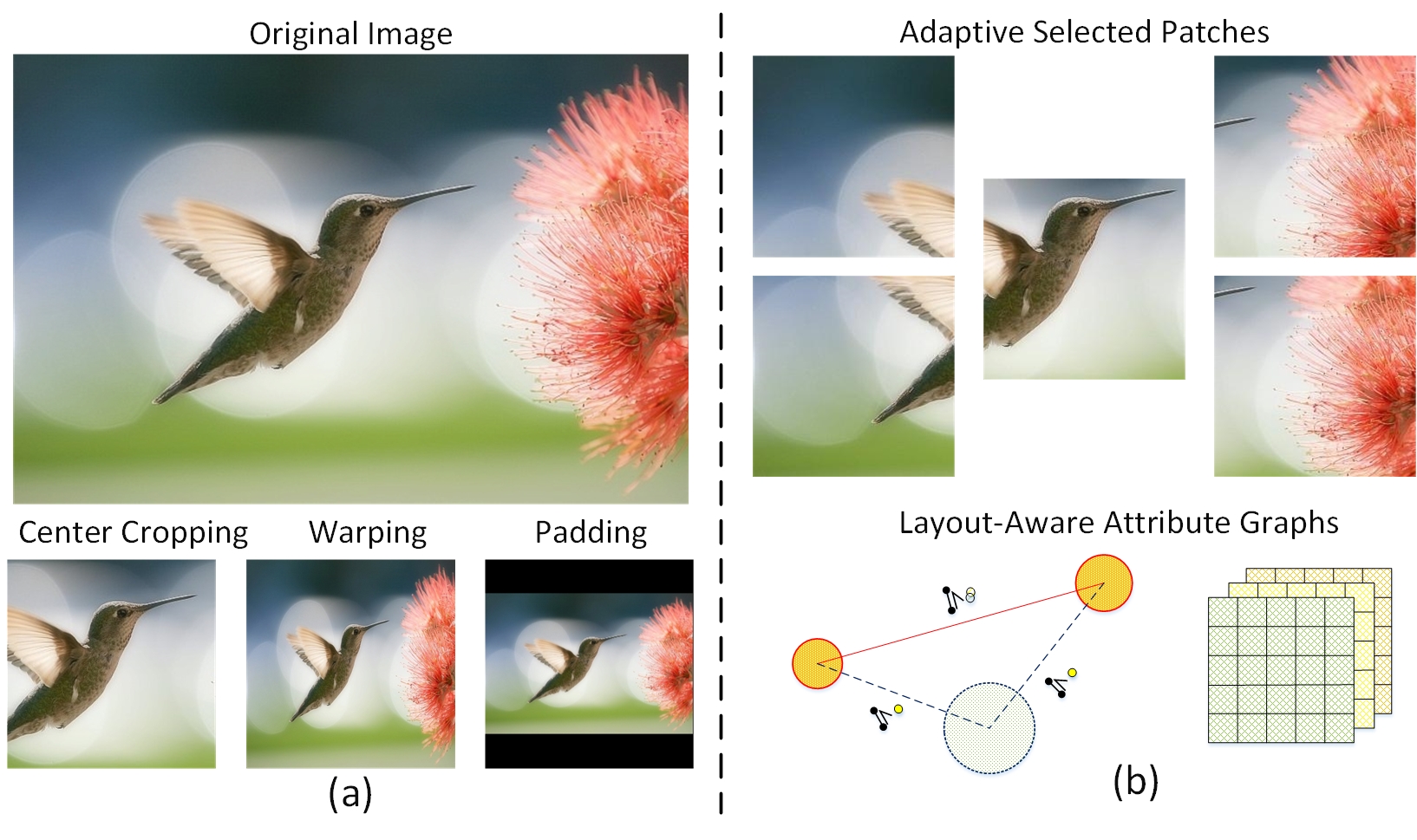}
		\label{transform}
		\caption{Conventional CNN methods (a) transform images via cropping, warping and padding. The proposed A-Lamp CNN (b) takes multiple patches and attributes graphs as inputs to represent fine grained details and the overall layout.}
		\vspace{-5mm}
	\end{figure}
	
	Automatic image aesthetics assessment is challenging. Among the early efforts \cite{Datta:2006:ECCV,Ke:2006:CVPR}, various hand-craft aesthetics features \cite{Luo:2008:ECCV,Bhattacharya:2010:ACMMM,Tang:2013:TMM,Sagnik:2011:CVPR,Su:2011:ACMMM,Cohen-Or:2006:SIGGRAPH} have been manually designed to approximate the photographic and psychological aesthetics rules. 
	However, to design effective aesthetics features manually is still a challenging task because even the very experienced photographers use very abstract terms to describe high quality photos.
	Other approaches leveraged more generic features\cite{Marchesotti:2011:ICCV,Perronnin:2010:ECCV,Su:2011:ACMMM} to predict photo aesthetics. However, these generic features may not be suitable for assessing photo aesthetics, as they are designed to capture the general characteristics of the natural images instead of describing the aesthetics of the images.
	
	Because of the limitations of these feature-based approaches, many researchers have recently turned to use deep learning strategy to extract effective aesthetics features \cite{Lu:2015:ICCV,Lu:2014:ACMMM,Mai:2016:CVPR,Tang:2014:CVPR,Karaye：2013：Archive:style}. These deep CNN methods have indeed shown promising results. However,
	the performance is often compromised by the constraint that the neural network only takes the fixed-size input. To accommodate this requirement, input images will need to be obtained via cropping, warping, or padding. As we can see from Figure \ref{intro}, these additional operations often alter image composition, reduce image resolution, or cause extra image distortion, and thus impair the aesthetics of the original images because of potential loss of fine grained details and holistic image layout. However, such fine-grained details and overall image layout are critical for the task of image quality assessment. He\cite{He:archive:2014} and Mai\cite{Mai:2016:CVPR} tried to address the limitation in fixed-size input by training images in a few different scales to mimic varied input sizes. However, they still learn from transformed images, which may result in substantial loss of fine grained details and undesired distortion of image layout. 
	
	Driven by this important issue, a question arises: \textit{Can we simultaneously learn fine-grained details and the overall layout to address the problems caused by the fixed-size limitation?} To resolve this technical issues, we present in this paper a dedicated CNN architecture named A-Lamp. This novel scheme can accept arbitrary images with its native size. Training and testing can be effectively performed by considering both fine-grained details and image layout, thus preserving the information from original images.
	
	Learning both fine-grained details and image layout is indeed very challenging.
	First, the detail information is contained in the original, high resolution images. Training deep networks with large-sized input dimensions requires much longer training time, training dataset, and hardware memory. To enable learning from fine grained details, a multi-patch-based method was proposed in \cite{Lu:2015:ICCV}. This scheme shows some promising results. However, these randomly picked bag of patches cannot represent the overall image layout. In addition, this random cropping strategy requires a large number of training epochs to cover the desired diversity in training, which lead to low efficiency in learning.
	
	Second, how to effectively describe specific image layout and incorporate it into the deep CNN is again very challenging. Existing works related to image layout descriptors are dominantly based on a few simple photography composition principles, such as visual balance, rule of thirds, golden ratio, and so on. However, these general photography principles are inadequate to represent local and global image layout variations. To incorporate global layout into CNN, transformed images via warping and center-cropping have been used to represent the global view \cite{Lu:2014:TMM:rating}.
	However, such transformation often alters the original image composition or causes undesired layout distortion. 
	
	\begin{figure}
		\centering
		\includegraphics[scale=0.23]{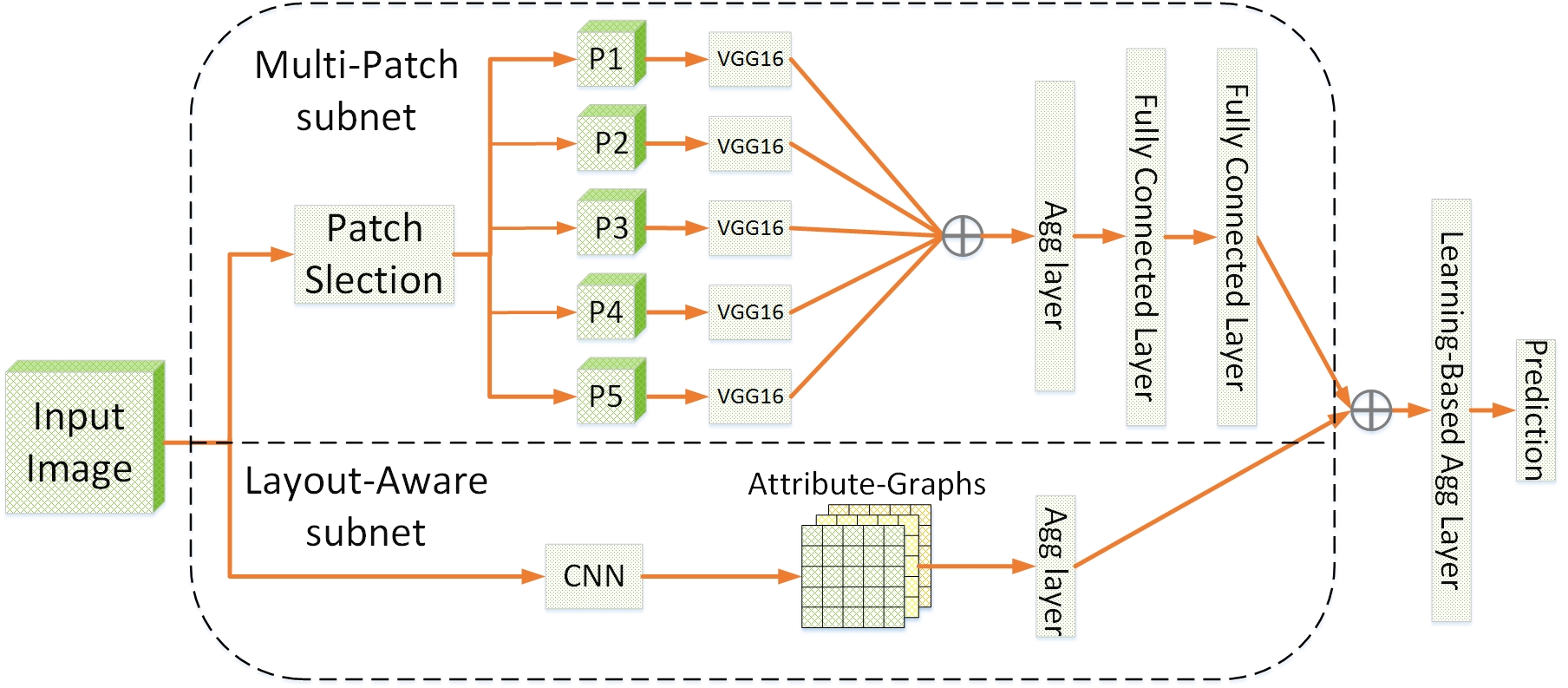}
		\caption{The architecture of the A-Lamp CNN. More detailed illustrations for Multi-Patch subnet and Layout-Aware subnet can be seen in Figure \ref{multi_patch} and Figure \ref{layout}.}
		\label{whole_net}
		\vspace{-3mm}
	\end{figure}
	
	In this paper, we resolved this challenges by developing an Adaptive Layout-Aware Multi-Patch Convolutional Neural Network (A-Lamp CNN) architecture. The design A-Lamp is inspired jointly by the success of fine-grained detail learning using multi-patch strategy \cite{Lu:2015:ICCV,Liu:2016:CVPR:multi_patch} and the success of holistic layout representation by attribute graph. However, the proposed scheme can successfully overcome the stringent limitations of the existing schemes.
	Like DMA-Net in \cite{Lu:2015:ICCV}, our proposed A-Lamp CNN also crops multiple patches from original images to preserve fine-grained details. Comparing to DMA-Net, this scheme has two major innovations. First, instead of cropping patches randomly, we propose an adaptive multi-patch selection strategy. The central idea of adaptive multi-patch selection is to maximize the input information more efficiently. We achieve this goal by dedicatedly selecting the patches that play important role in affecting images' aesthetics. We expect that the proposed strategy shall be able to outperform the random cropping scheme even with substantially less training epochs. Second, unlike the DMA-Net that just focus on the fine-grained details, this A-Lamp CNN incorporates the holistic layout via the construction of attribute graph. We use graph nodes to represent objects and the global scene in the image. Each object (note) is described using object-specific local attributes while the overall scene is represented with global attributes. The combination of both local and global attributes captures the layout of an image effectively. This attribute-graphs based approach is expected to model image layout more accurately and outperform the existing approaches based on warping and center-cropping. These two innovations result improvement in both efficiency and accuracy over DMA-Net. The main contributions of this proposed A-Lamp scheme can be summarized into three-fold:
	
	
	$\bullet$ We introduce a new neural network architecture to support learning from any image sizes without being limited to small and fixed size of the image input. This shall open a new avenue of deep learning research on arbitrary image size for training.
	
	$\bullet$ We design two novel subnets to support learning at different levels of information extraction: fine-grained image details and holistic image layout. Aggregation strategy is developed to effectively combine hybrid information extracted from individual subnet learning.
	
	$\bullet$ We have also developed an adaptive patch selection strategy to enhance the training efficiency associated with variable size images being used as the input. This aesthetics driven selection strategy can be extended to other image analysis tasks with clearly-defined objectives.
	
	\section{Related Work}
	\subsection{Deep Convolutional Neural Networks}
	Deep learning methods have shown great successes in various 
	computer vision tasks, including conventional tasks in object recognition \cite{Xu:2016:CVPR:object_selection}, object detection \cite{He:archive:2014,Liu:2016:CVPR:multi_patch}, and image classification \cite{Reed:2016:CVPR:classification,He:2016:CVPR:residual}, as well as contemporary tasks in image captioning \cite{Hendricks:2016:CVPR:captioning}, saliency detection \cite{Pan:2016:CVPR:saliency}, style recognition \cite{Gatys:2016:CVPR:style_transfer,Karaye：2013：Archive:style} and photo aesthetics assessment \cite{Lu:2014:ACMMM,Lu:2015:ICCV,Tang:2014:CVPR,Mai:2016:CVPR,Kang:2014:CVPR}. Most existing deep learning methods transform input images via cropping, scaling, and padding to accommodate the deep neural network architecture requirement in fixed size input which would compromise the network performance as we have discussed previously.
	
	Recently, new strategies to construct adaptive spatial pooling layers have been proposed to alleviate the fixed-size restriction \cite{He:archive:2014,Mai:2016:CVPR}. In theory, these network structures can be trained with standard back-propagation, regardless of the input image size. In practice, the GPU implementations of deep learning are preferably run on fixed input size. The recent research \cite{He:archive:2014,Mai:2016:CVPR} mimic the variable input sizes by using multiple fixed-size inputs which are obtained by scaling from original images. It is apparently still far from arbitrary size input. Moreover, the learning is still from transformed images, which inherently compromise the performance of the deep learning networks.
	
	Others have proposed dedicated network architectures. A double-column deep convolutional neural network was developed in \cite{Lu:2014:ACMMM} to support heterogeneous inputs with both global and local views. The global view is represented by padded or warped image while, the local view is represented by randomly cropped single patch. This work was further improved in \cite{Lu:2015:ICCV}, where a deep multi-patch aggregation network was developed (DMA-Net) to take multiple randomly cropped patches as input. This network have shown some promising results. However, these random order of bag of patches is unable to capture image layout information, which is crucial in image aesthetics assessment. Furthermore, to ensure that most of the information will be captured by the network, this scheme uses large number of randomly selected groups of patches for each image, and trains them for 50 epochs, resulting in very low training efficiency.
	
	\subsection{Image Layout Representation}
	To represent holistic image layout, existing works \cite{2010:optimizing_composition,2010:ICIP:composition,2012:ICIP:composition,ICML:2012:composition,Yao:2012:oscar,Ma:2014:ICIP:photography,Ma:2014:ACMMM:pose,Zhou:2015:ACMMM:vanishing_point} adopt dominantly the model of image composition by approximating some simple traditional photography composition guidelines, such as visual balance, rule of thirds, golden ratio, and diagonal dominance. However, these heuristic guidance-based descriptors cannot capture the intrinsic of photo aesthetics in terms of image layout. 
	
	Attribute-graph, which has long been used by the vision community to represent structured groups of objects \cite{Felzenszwalb:2004:graph_seg,Lu:2014:graph,CVPR:2014:multigraph,Shi:2000:PAMI:graph,TIP:2014:graphlet}, shows promising results in representing complicated image layout.   The spatial relationship between a pair of objects was considered in \cite{Lan:2012:ECCV:retrieval} even though the overall geometrical layout of all the objects and the object characteristics cannot be accounted for with this method. The scheme reported in \cite{Xu:2010:SIGAPH:concep_map} was able to maintain spatial relationships among objects but related background information and object attributes were not addressed. The scheme reported in  \cite{2013:PAMI:description} considers both objects and their interrelations, but have not been integrated with the holistic background modeling. The scheme in \cite{Cao:2014:ACMMM:layout} performs image aesthetics ranking by constructing the triangular object structures with attribute features. However, this scheme lacks of proper account for the global scene context.
	
	\begin{figure}
		\centering
		\includegraphics[scale=0.27]{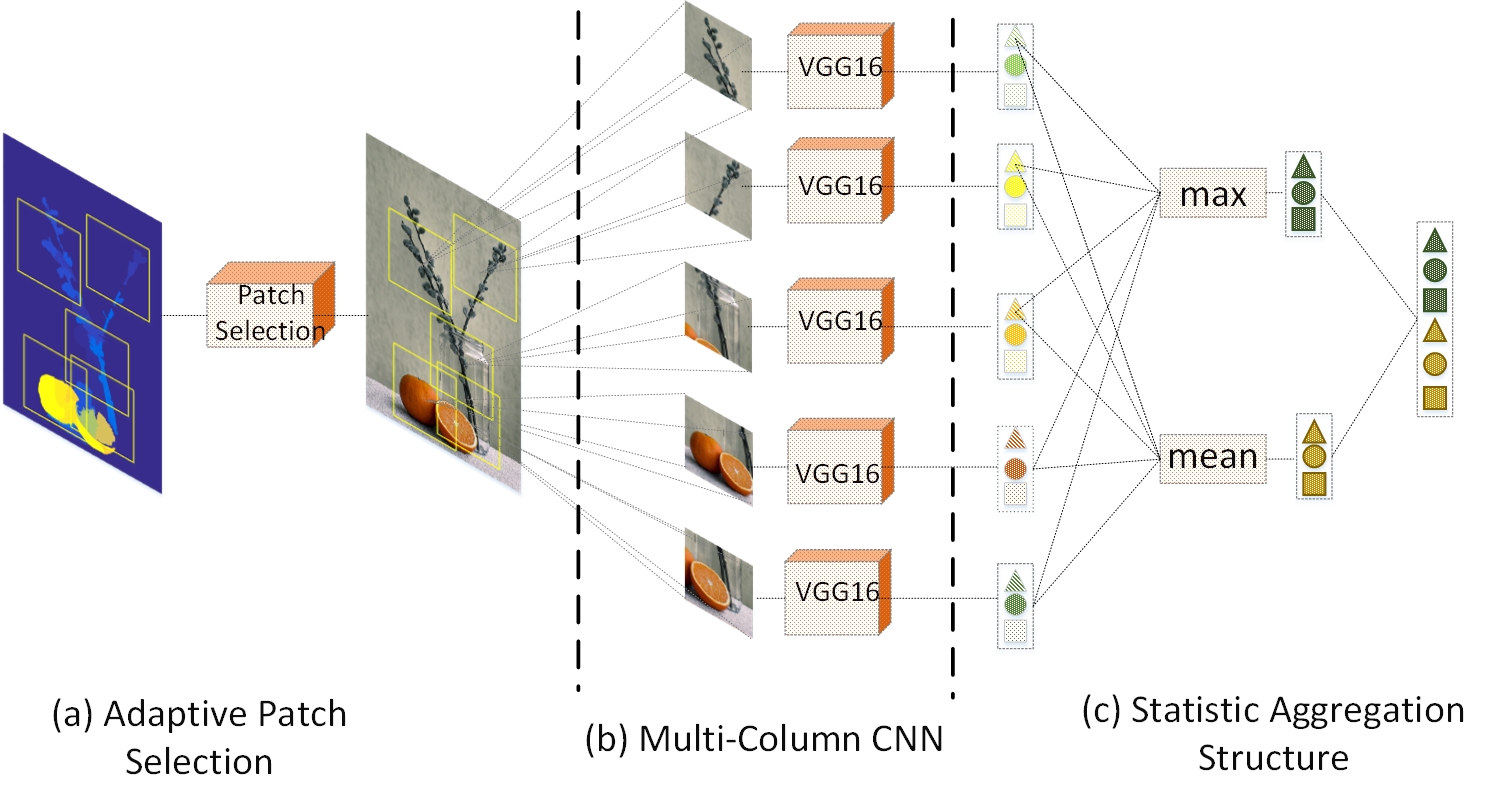}
		\caption{The architecture of Multi-Patch subnet: (a) adaptive patch selection module, (b) a set of paralleled shared weights CNNs that are used for extracting deep features from each of the patch, (c) aggregation structure which combines the extracted deep features from the multi-column CNNs jointly.}
		\label{multi_patch}
		\vspace{-3mm}
	\end{figure} 
	\section{Adaptive Layout-Aware Multi-Patch CNN}
	The architecture of the proposed A-Lamp is shown in Figure \ref{whole_net}. Given an arbitrary sized image, multiple patches will be adaptively selected by the \textit{Patch Selection} module, and fed into the Multi-Patch subnet. A statistic aggregation layer is followed to effectively combine the extracted features from these multiple channels.
	At the same time, a trained CNN is adopted to detect salient objects in the image. The local and global layout of the input image are further represented by \textit{Attribute-Graphs}. At the end, a learning-based aggregation layer is utilized to incorporate the hybrid features from the two subnets and finally produce the aesthetic prediction. More details will be illustrated in this section. 
	\begin{figure*}
		\centering
		\includegraphics[scale=0.4]{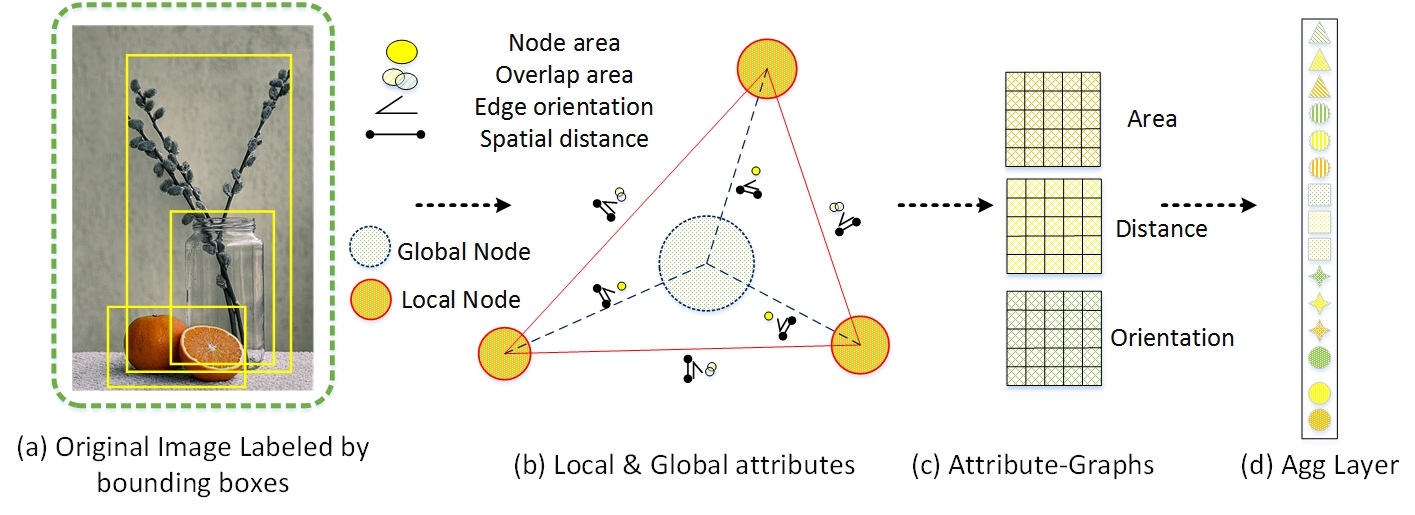}
		\caption{Pipeline of attribute-graphs construction. (a) Salient objects (labeled by yellow bounding boxes) are first detected by a trained CNN, and regarded as local nodes. The dashed green bounding box denote the overall scene, which served as a global node. (b) Local and global attributes are extracted from these nodes to capture the object topology and the image layout. (c) Attribute-graphs are constructed and (d) concatenated into an aggregation layer.}
		\label{layout}
		\vspace{-3mm}
	\end{figure*}
	
	\subsection{Multi-Patch subnet}
	We represent each image with a set of carefully cropped patches, and associate the set with the image's label.
	The training data is ${{\rm{\{ }}{{\rm{P}}_n}{\rm{, }}{{\rm{y}}_n}{\rm{\} }}_{n \in [1,N]}}$, where ${P_n} = {\{ {p_{nm}}\} _{m \in [1,M]}}$ is the set of $M$ patches cropped from each image. The architecture of proposed Multi-Patch subnet is shown in Figure \ref{multi_patch} and more details will be explained in this section. 
	
	\subsubsection{Adaptive Patch Selection}
	Different from the random-cropping method in \cite{Lu:2015:ICCV}, we aim to carefully select the most discriminative and informative patches to enhance the training efficiency. To realize that, we studied professional photography rules and human visual principles. It has been observed that, human visual attention does not distribute evenly within an image. That means some regions play more important roles than other regions when people viewing photos. In addition, holistic analysis is critical for evaluating an image's aesthetics. It has been shown that focusing on the subjects is often not enough for overall aesthetic assessment. Motivated by these observations, several criteria have been developed to perform patch selection:
	
	\textbf{Saliency Map}. The task of saliency detection is to identify the most important and informative part of a scene. Saliency map models human visual attention, and is capable of highlighting visually significant region. Therefore, it is natural to adopt saliency map for selecting regions that human usually pay more attention to.
	
	\textbf{Pattern Diversity}. In addition to saliency map, we also encourage diversification within a set of patches. Different from conventional computer vision tasks, such as image classification and object recognition, that often focus on the foreground objects, image aesthetics assessment also heavily depends on holistic analysis of entire scene. Important aesthetic characteristics, e.g. Low-of-Depth, color harmonization and simplicity, can only be perceived by analyzing both the foreground and background as a whole. 
	
	\textbf{Overlapping Constraint}. Spatial distance among any patch pairs should also be considered to constrain the overlapped ratio of these selected patches. 
	
	Therefore, we can formulate the patch selection as an optimization problem.  An objective function can be defined to search for the optimal combination of patches:
	\begin{equation}
	\{ {c^*}\}  = \mathop {argmax}\limits_{i,j \in [1,M]} F\left( {S,\;{D_p},{D_s}} \right)
	\end{equation}
	\begin{equation}
	F( \cdot ) = \sum\limits_{i = 1}^M {{S_i}}  + \sum\limits_{i \ne j}^M {{D_p}({{{\rm{\tilde N}}}_i},{{{\rm{\tilde N}}}_j}) + \sum\limits_{i \ne j}^M {{D_s}({c_i},{c_j})} }
	\end{equation}
	where ${\{ c_m^ * \} _{m \in [1,M]}}$ is the centers of the optimal set of $M$ selected patches. $\;{S_i} = \frac{{sal({p_i})}}{{area({p_i})}}$ is the normalized saliency value for 
	each patch $p_i$. The saliency value is obtained by a graph-based saliency detection approach \cite{C.Yang:2013:CVPR:graph_based_saliency}.
	${D_p}( \cdot )$ is the pattern distance function which 
	measures the difference of two patches' patterns. 
	Here we adopt edge and chrominance distribution to represent the pattern of each patch. Specifically, we model the pattern of a patch $p_m$ using a multi-variant Gaussian:
	\begin{equation}
	{\tilde N_m} = {\{ {\{ {N_e}({\mu _e},{\Sigma _e})\} _m},{\{ {N_c}({\mu _c},{\Sigma _c})\} _m}\} _{m \in [1,M]}}
	\end{equation}
	where ${\{ {N_e}({\mu _e},{\Sigma e})\} _m}$ and ${\{ {N_c}({\mu _c},{\Sigma _c})\} _m}$ denote edge distribution and chrominance distribution of patch $p_m$, respectively. ${\Sigma _e}$ and ${\Sigma _c}$ are the covariance matrices of $N_e$ and $N_c$. Therefore, measuring pattern difference between a pair of patches can be formulated by mapping these distributions $\tilde N_m$ to the \textit{Wasserstein Metric space} $M_{m \times m}$,  and calculate the \textit{$1_{st}$ Wasserstein distance} between ${\tilde N_i}$ and ${\tilde N_j}$ on this given metric space $M$.
	Following F. Pitie \cite{F.Pitie:2007:color_transfer}, the closed form solution is given by: 
	\begin{equation}
	{D_p}( \cdot ) = \Sigma _i^{ - 1/2}\left( {\Sigma _i^{1/2}{\Sigma _j}\Sigma _i^{1/2}} \right)\Sigma _i^{ - 1/2}
	\end{equation}
	${D_s}( \cdot )$ is the spatial distance function, which is measured by Euclidean Distance.
	
	\subsubsection{Orderless Aggregation Structure}
	We also perform the aggregation of the multiple instances to enable the proposed network learn from multiple patches cropped from a given image. Let ${\left\langle {{\rm{Blo}}{{\rm{b}}_n}} \right\rangle _{l}} = \{ b_i^n\} _{i \in [1,M]}^{l}$ be the set of patch features extracted from $n_{th}$ image 
	at $l_{th}$ layer of the shared CNNs. $b_{i,l}^n$ is a $K$ dimensional vector.
	$T_k$ denotes the set of values of the $k_{th}$ component of all $b_{i,l}^n \in {\left\langle {Blo{b_n}} \right\rangle _l}$. For simplicity, we omit image index $n$ and layer index $l$, thus ${T_k} = {\{ {d_{ik}}\} _{i \in [1,M]}}$.
	The aggregation layer is comprised of a collection of statistical functions, 
	i.e., ${F_{Agg}} = {\{ F_{Agg}^u\} _{u \in [1,U]}}$. Each $F_{Agg}^u$ computes
	$Blob$ returned by the shared CNNs. Here we adopt a modified statistical functions proposed in \cite{Lu:2015:ICCV}, i.e. $U = \{ max,\;mean\} $\footnote[1]{Through extensive experiments, we find that {max, min} showing the best performance. The statistical functions adopted in \cite{Lu:2015:ICCV}, i.e. {min, max, mean, median}, not result in performance improvement, and even worse because the potential of over-fitting caused by the too large vector dimension.}. The outputs of the functions in $U$ are concatenated to produce a ${K_{stat}}$-dimensional feature vectors. Two fully connected layers are followed to implement multi-patch
	aggregation component. The whole structure can be expressed as a function 
	$f:\{ Blob\}  \to {{{K_{stat}}}}$:
	\begin{equation}
	f(Blob) = W \times ( \oplus _{u = 1}^U \oplus _{k = 1}^KF_{Agg}^u({T_k}))
	\end{equation}
	where $ \oplus$ is a vector concatenation operator which produces a column vector, $W \in {^{{K_{stat}} \times UK}}$ is the parameters of the fully-connected layer. 
	Figure \ref{multi_patch} shows an example of Statistics Aggregation Structure with $M = 5$ and $K=3$. In practice, the feature dimension $K=4096$.
	
	\begin{figure}
		\centering
		\includegraphics[scale=0.11]{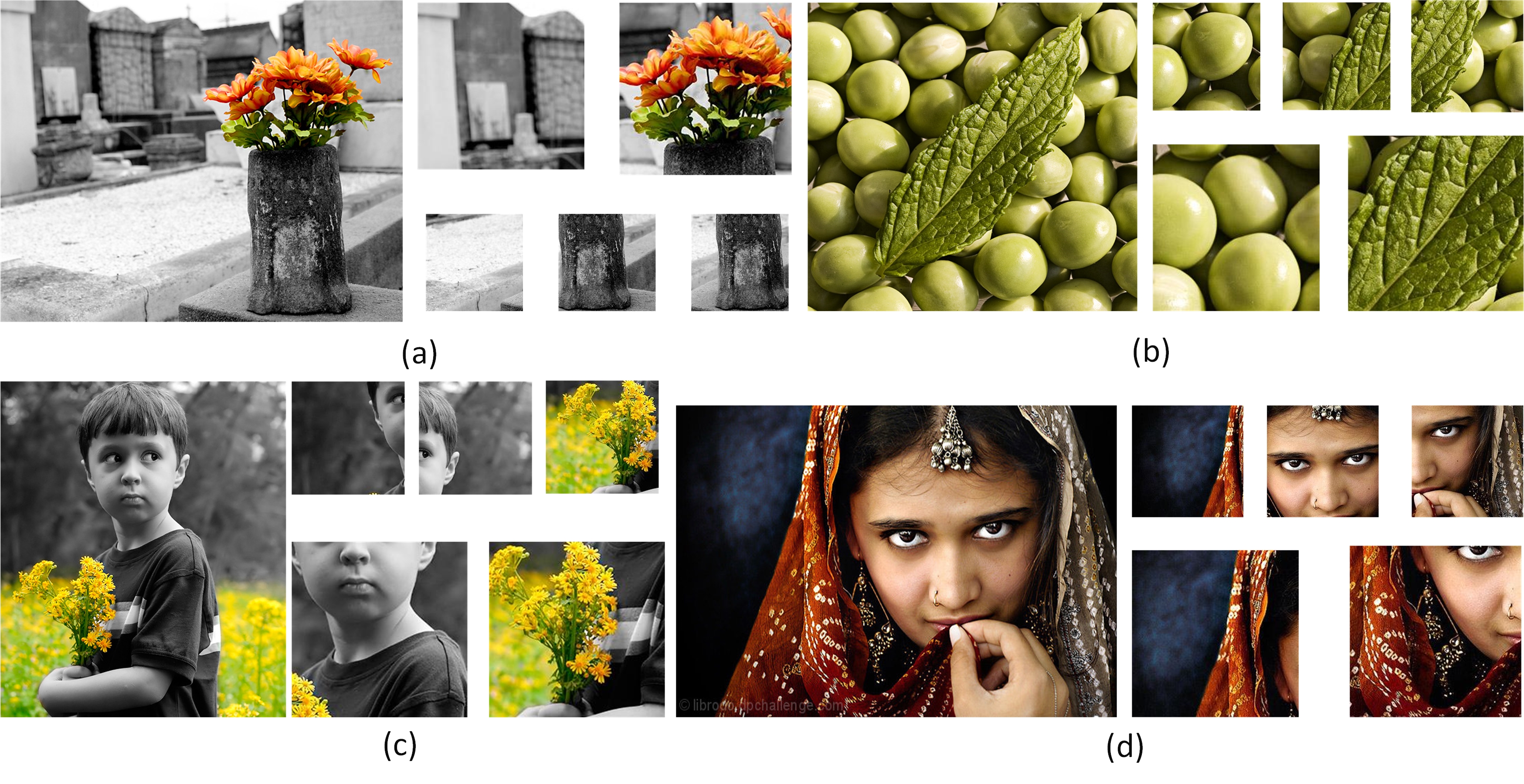}
		\caption{Examples of selected patches by the proposed Adaptive Patch-Selection scheme. In each group, original image is on the left side, and patches are located on the right side. We zoom in the patches that have more details for clear display. In practice, the size of the all the patches are 224 $\times$ 224. }
		\label{patches}
		\vspace{-3mm}
	\end{figure}
	\subsection{Layout-Aware Subnet}
	We first employ a trained CNN  \cite{zhang:2015:CVPR:boundingbox} to localize the salient objects. Let ${\rm I}:{\{ {{\rm B}_i},{s_i}\} _{{N_{obj}}}}$ denotes a set of detected objects in image ${\rm I}$, where each object is labeled by a bounding box ${{\rm B}_i}$ and associated with a confidence score ${s_i}$, $N_{obj}$ denotes the number of objects.
	Here $G(V,E)$ is an undirected fully connected graph. 
	$V$ represents the nodes and $E$ represents the set of edges connecting the nodes. We define two types of attributes in this research:
	
	\textbf{Local Attributes}. Each object presents in the image contributes to a graph node resulting in a total of $N_{obj}$ local nodes ${V_{l}} = \{ {v_1}, \cdot  \cdot  \cdot ,{v_{{N_{obj}}}}\} $. local edges $E_{l}$ refer to the edges between a pair of local nodes, there will be $({N_{obj}} - 1)!$ such edges. Each local node is represented using local attributes. These local attributes are limited to the area occupied by the bounding box of that particular object. The local attributes capture the relative arrangement of the objects with respect to each other, which are represented by 
	\begin{equation}
	{\Phi _{l}}(i,j) = {\{ dist(i,j),\;\theta (i,j),\;\hat o(i,j)\} _{{v_i},{v_j} \in {V_{l}}}}
	\end{equation}
	where ${\Phi _{l}}(i,j)$ represents the attribute of a pair of connecting node $v_i$ and $v_j$. $dist(i,j)$ is the spatial distance between object centroids. $\theta(i,j)$ represents the angle of the graph edge with respect to the horizontal taken in the anti-clockwise direction. It indicates the relative spatial organization of the two objects. $\hat o(i,j)$ represents the amount of overlap between the bounding boxes of the two objects and is given by 
	\vspace{-3mm}
	\begin{equation}
	{\hat o_{ij}} = \frac{{area({v_i}) \cap area({v_j})}}{{\min (area({v_i}),area({v_j}))}}
	\end{equation}
	where ${area({v_i})}$ is the fraction of the image area occupied by the ${i^{th}}$ bounding box. The intersection of the two bounding boxes is normalized by the smaller of the bounding boxes to ensure the overlap score of one, when a smaller object is inside a larger one.
	
	\textbf{Global Attributes}. The global node $V_{g}$ represents the overall scene. The edges connecting local nodes and global node are global edges $E_{g}$, there will be $N_{obj}$ such edges.
	The global node captures the overall essence of the image. The global attributes ${\Phi_{g}}$ are given by
	\begin{equation}
	{\Phi _g}(i,g) = {\{ dist(i,g),\;\theta (i,g),\;area({v_i})\} _{{v_i} \in {V_l},{v_g} \in {V_g}}}
	\end{equation}
	where $dist(i,g)$ and $\theta (i,g)$ are the magnitude and orientation of the edge connecting the centroid of the object corresponding to node $v_i$ to the global centroid $c_g$. The edges connecting each object to the global node illustrate the placement of that object with respect to the overall object topology. 
	
	An aggregation layer is adopted to concatenate the constructed attribute graphs into a feature vector ${\vec \nu }$, and further combined with the Multi-Patch subnet, which is illustrated in Figure \ref{whole_net}.\footnote[2]{By statistical study, we find that, the confidence score is very low when ${N_{obj}} \ge 5$. So we set $N_{obj} = 4$ to fix the feature vector ${\vec \nu }$ dimension.} 
	
	\section{Experimental Results} \label{experiments}
	In the implementation, we release the memory burden by first training the Multi-Patch subnet and then combining with the Layout-Aware subnet to fine-tune the overall A-Lamp. The weights of multiple shared column CNNs in the Multi-Patch subnet are initialized by the weights of VGG16. VGG16 is one of the state-of-the-art object-recognition networks that is pre-trained on the ImageNet \cite{Alex:2012:NIPS:ImageNet}.  
	Following Lu \cite{Lu:2015:ICCV}, The number of patches in a bag is set to be 5. The patch size is fixed to be 224 $ \times 224 \times 3$. The base learning rate is 0.01, the weight decay is 1e-5 and momentum is 0.9.  
	All the network training and testing are done by using the Caffe deep learning framework\cite{Jia:2014:ACMMM:caffe}. 
	
	We systematically evaluate the proposed scheme on the AVA dataset \cite{Murray:MMP2012:AVA}, which, to our best knowledge, is the largest publicly available aesthetic assessment dataset. The AVA dataset provides about 250,000 images in total. The aesthetics quality of each image in the dataset was rated on average by roughly 200 people with the ratings ranging from one to ten, with ten indicating the highest aesthetics quality. For a fair comparison, we use the same partition of training data and testing data as the previous
	work \cite{Lu:2014:ACMMM,Lu:2015:ICCV,Mai:2016:CVPR,Murray:MMP2012:AVA} in which roughly 20,0000 images are used for training and 19,000 images for testing. We also follow the same procedure as previous works to assign a binary aesthetics label to each image in the benchmark. Specifically, images with mean ratings smaller or equal to 5 are labeled as low quality and those with mean ratings larger than 5 are labeled as high quality.
	
	\begin{table}
		\begin{center}	
			\begin{tabular}{||c|c||} 
				\hline
				Method & Accuracy \\
				\hline\hline
				DMA-Net{\tiny ave} & 73.1 $\%$ \\
				DMA-Net{\tiny max} & 73.9 $\%$ \\
				DMA-Net{\tiny stat} & 75.4$\% $ \\
				DMA-Net{\tiny fc} & 75.4$\% $ \\
				\hline
				Random-MP-Net & 74.8$\%$ \\
				\textbf{New-MP-Net} & \textbf{81.7$\% $}\\
				\hline
			\end{tabular} 
		\end{center} 
		\caption{Performance comparisons of Adaptive Multi-Patch subnet with other multi-patch-based CNNs.}
		\label{tabel1} 
		\vspace{-3mm}
	\end{table}
	
	\subsection{Analysis of Adaptive Multi-Patch subnet}
	\begin{table}
		\begin{center}
			\begin{tabular}{||c|c|c||} 
				\hline
				Method & Accuracy & F-measure\\
				\hline\hline
				AVA & 67.0 $\%$ & $N{A^ * }$  \\
				VGG-Center-Crop & 72.2 $\%$ & 0.83 \\
				VGG-Wrap & 74.1 $\%$ & 0.84 \\
				VGG-Pad & 72.9 $\%$ & 0.83 \\
				\hline
				SPP-CNN & 76.0 $\%$ & 0.84 \\
				MNA-CNN & 77.1 $\%$ & 0.85 \\
				MNA-CNN-Scene & 77.4 $\%$ & $N{A^ * }$ \\
				DCNN & 73.25 $\%$ & $N{A^ * }$ \\
				DMA-Net-ImgFu & 75.4 $\%$ & $N{A^ * }$\\
				\hline\hline
				\textbf{New-MP-Net} & \textbf{81.7$\% $} & \textbf{0.91} \\
				\textbf{A-Lamp} & \textbf{82.5 $\%$} & \textbf{0.92} \\
				\hline
			\end{tabular} 
		\end{center}
		\caption{Comparisons of A-lamp with the state-of-the-art. $^*$ These results are not reported in the original papers \cite{Lu:2015:ICCV,Lu:2014:TMM:rating,Mai:2016:CVPR,Murray:MMP2012:AVA}.}
		\label{tabel2}
		\vspace{-3mm}
	\end{table} 
	For a fair comparison, we first perform the training and testing only using the proposed Multi-Patch subnet, and evaluate it with some other multi-patch-based networks. 
	
	\textbf{DMA-Net}. DMA-Net proposed in \cite{Lu:2015:ICCV} is a very recent dedicated deep Multi-Patch-based CNN for aesthetic assessment.
	Specifically, DMA-Net performs multi-column CNN training and testing. Five randomly cropped patches from each image was used as training, and the label of the image is associated with the bag of patches. Here we compare New-MP-Net with four types of DMA-Net architecture. DMA-Net{\tiny ave} and DMA-Net{\tiny max} train the DMA-Net using standard patch pooling scheme, where DMA-Net{\tiny ave} performs average pooling and DMA-Net{\tiny max} performs max pooling. The DMA-Net using Statistics Aggregation Structure is denoted as DMA-Net{\tiny stat} and Fully-Connected Sorting Aggregation Structure as DMA-Net{\tiny fc}. 
	
	\textbf{MP-Net}. The Multi-Patch subnet that takes the inputs by our proposed adaptive patch selection scheme is denoted as \textbf{New-MP-Net}. Since we adopt much deeper shared column CNNs (VGG16)  in \textbf{New-MP-Net}. One may argue that the better performance may rely on the adoption of VGG16. Therefore, we train and test New-MP-Net by the same random cropping strategy in \cite{Lu:2015:ICCV}, which is denoted as Random-MP-Net. Specifically, we randomly crop 50 groups of patches from the original image with a 224$\times$224 cropping window. For each testing image, we perform prediction for 50 random crops and take their average as the final prediction result. 
	
	The experimental results are shown in Table \ref{tabel1}. We can see that, New-MP-Net outperforms all types of DMA-Net architectures.
	Although DMA-Net randomly cropped 50 groups of patches to train, and the total training has 50 epochs. The randomness in cropping was not able to effectively capture useful information and may cause the training to be confusing for the network. Besides, we find that most of the random generated patches are cropped from the same location of the image. That means, there are a large number of data repeatedly fed into the network, thus lead to the risk of over-fitting. Comparing the accuracy and F-measure of New-MP-Net (81.7$\%$ and 0.91) with Random-MP-Net (71.2$\%$ and F-measure 0.83), we can see that even using the same network architecture, the performance is impaired by using random-cropping strategy.
	
	\subsection{Effectiveness of Adaptive Patch Selection}
	Instead of random cropping, we adaptively select the most informative and discriminative patches as input, which is the key to achieve substantial performance enhancement. From Figure \ref{intro}, we can see that, the salient objects, i.e. the bird and the flower, have been selected. Within these patches, the most important information and the fine-grained details are all retained. In addition, the background which shows different patterns, i.e. the blue sky and the green ground, have also been selected. Therefore, the global characteristics, e.g. color harmony, Low-of-Depth, can also be perceived by learning these patches jointly. More examples of selected patches are shown in Figure \ref{patches}. We can see that, the proposed adaptive selection strategy not only is effective in selecting the most salient regions (e.g. the human's eyes, face and the orange flowers), but also is capable of capturing the pattern diversity (e.g. the green leaf and green beans, the flower and the gray wall). Furthermore, the proposed adaptive patch selection strategy is also able to enhance the training efficiency. The result of New-MP-Net is obtained by taking 20-30 training epochs, substantially less than 50 epochs reported in \cite{Lu:2015:ICCV}, while still achieving  better performance. 
	
	\subsection{Comparison with the State-of-the-Art}
	Table \ref{tabel2} shows the results of the proposed A-Lamp CNN on the AVA dataset \cite{Murray:MMP2012:AVA} for image aesthetics categorization. The AVA dataset provides the state-of-the-art results for methods that use manually designed features and generic image features for aesthetics assessment. It is obvious that, all recently developed deep CNN schemes outperform these conventional feature-based approaches.
	
	\textbf{A-Lamp vs. Baseline}.
	To examine the effectiveness of the proposed scheme, we compare New-MP-Net and A-Lamp with some baseline methods that take only fixed-size inputs. In particular, we experiment on VGG16 with three types of transformed inputs.
	The input of VGG16-Center-Crop is obtained by cropping  from the center of the original image with a 224$\times$224 cropping window. The input of VGG16-Wrap is obtained by scaling the original input image to the fixed size of 224$\times$224. In the experiment of VGG16-Pad, the original image is uniformly resized such that the larger dimension becomes 224 and the aspect ratio is preserved. The 224$\times$224 input is then formed by padding the remaining dimension of the transformed image with zero-valued pixels.
	We can see from Table \ref{tabel2} that, both New-MP-Net and A-Lamp outperform these fixed-size input VGG nets. Such results confirmed that training network on multiple patches produces better prediction than networks training on a single patch.
	
	\textbf{A-Lamp vs. Non-fixed-Size CNNs}.
	We also compared the proposed scheme with some latest non-fixed size restriction schemes, i.e. SPP-CNN \cite{He:archive:2014} and MNA-CNN \cite{Mai:2016:CVPR}. Different from these schemes that their inputs are from several different level of scaled images, we implement the A-Lamp network to be trained from the original images. The results confirm that learning from original images is essential for aesthetic assessment, as we have discussed earlier. In addition, higher prediction accuracy of the proposed scheme further proves that, the adaptive Multi-Patch strategy is more efficient than the spatial pyramid pooling strategy adopted in SPP-CNN and MNA-CNN.
	
	\textbf{A-Lamp vs. Layout-Aware CNNs }.
	
	\romannum{1}. To show the effectiveness of the proposed layout-aware subnet, we compare A-Lamp with several latest deep CNN networks that incorporate global information for learning. MNA-CNN-Scene \cite{Mai:2016:CVPR} replace the average operator in the MNA-CNN network with a new aggregation layer that takes the concatenation of the sub-network predictions and the image scene categorization posteriors as input to produce the final aesthetics prediction. We can see from the results that incorporating scene attributes does not cause noticeable performance improvement.
	
	\begin{figure}
		\centering
		\includegraphics[scale=0.18]{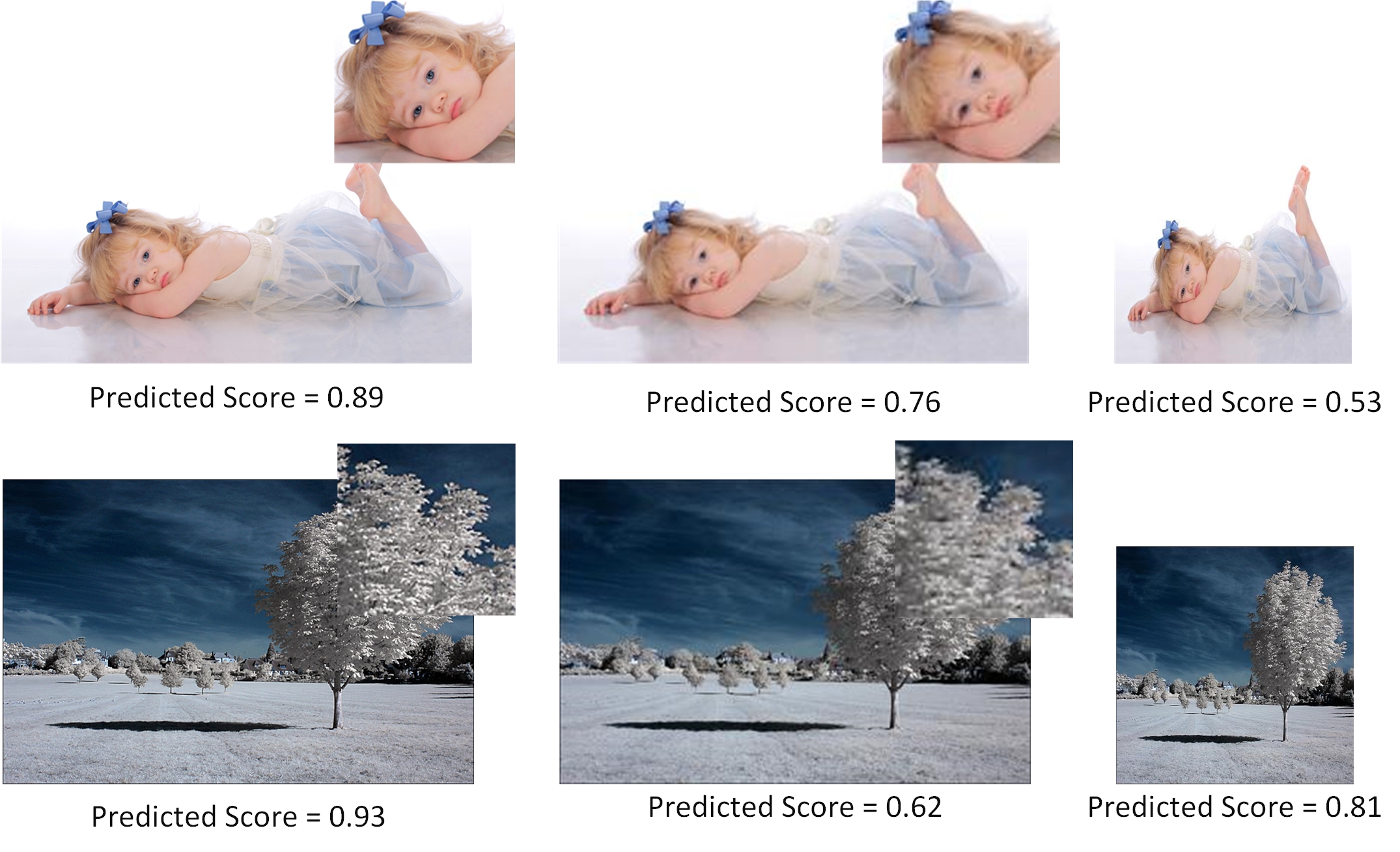}
		\caption{Prediction results on transformed images. Images from left to right are original ones, down sampled version and warped version. We zoom in some regions for comparison the details of original images and the down sampled images }
		\label{trans_img}	
		\vspace{-3mm}
	\end{figure}

	\begin{figure*}
		\centering
		\includegraphics[scale=0.25]{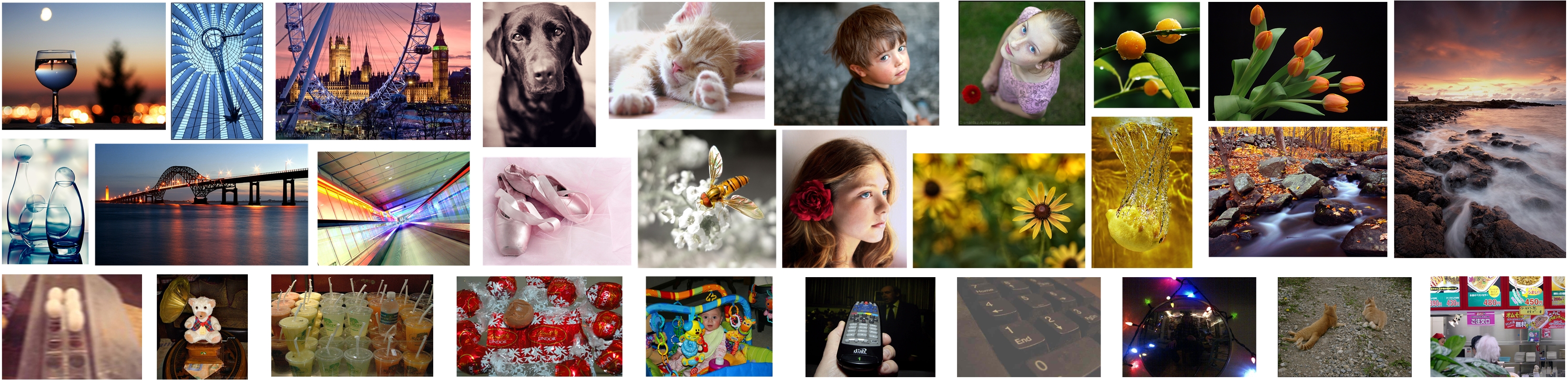}
		\caption{Results of predicted photos. The top two rows are predicted photos with high aesthetic scores. We random select these photos from eight categories \cite{Murray:MMP2012:AVA}. The low aesthetic quality photos are shown in the third row.}
		\label{samples}
		\vspace{-1mm}
	\end{figure*}
	
	\begin{figure} [!t]
		\centering
		\includegraphics[width=7.8cm, height=3.5cm]{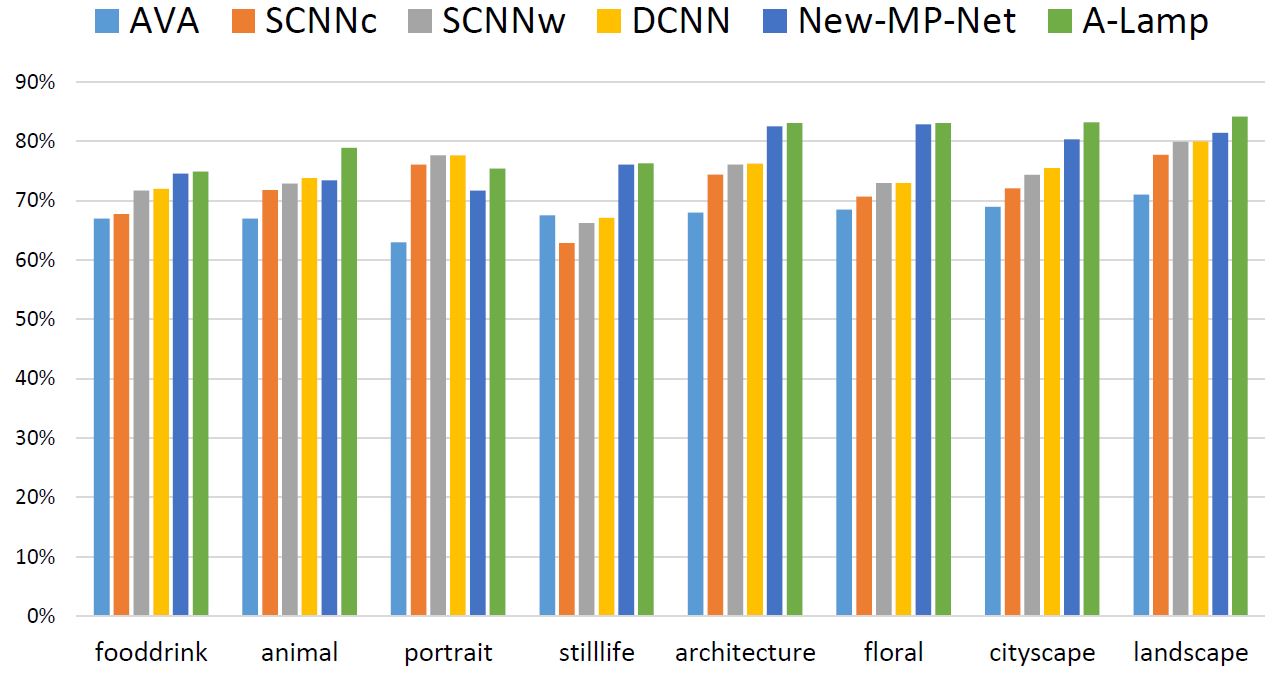}
		\caption{Comparison of aesthetic prediction performance in different content-based categories.}
		\label{category}
		\vspace{-5mm}
	\end{figure}
	
	\romannum{2}. DCNN \cite{Lu:2014:TMM:rating} is a double column convolutional neural network which combines random cropped and warped images as inputs to perform training . By comparing the test accuracy of the proposed A-Lamp (82.5 $\%$) with that of DCNN (73.25 $\%$), we can conclude that using randomly cropped and warped images to capture local and global image characters is not as effective as our approach.
	
	\romannum{3}. The result of DMA-Net-ImgFu (75.4 $\%$) \cite{Lu:2015:ICCV} is obtained by averaging the prediction results of DMA-Net and the fine tuned Alexnet \cite{Alex:2012:NIPS:ImageNet}. It is interesting that, though they incorporated transformed entire images to represent global information, it still fall behind the performance of our proposed A-Lamp (82.5 $\%$). Such results further validate the effectiveness of our proposed layout-aware subnet.

	\subsection{A-Lamp Effectiveness Analysis}
	From Table \ref{tabel2}, we can see that, the proposed layout-aware approach boosts the performance of New-MP-Net slightly, but outperforms significantly over the other state-of-the-art approaches. The overall results show that both holistic layout information and fine-grained information are essential for image aesthetics categorization. 
	
	We further examined whether or not the proposed A-Lamp network is capable of responding to the changes in image holistic layout and fine grained details. To test this, we random collect 20 high quality images from the AVA dataset. We generate a down sampled version and a warped version from the original image. The down-sampled version keeps the same aspect ratio (i.e. the layout has not be changed) but reduced to one half of the original dimension. The warped version is generated by scaling along the longer edge to make it square. From the predicted aesthetic score we can confirm that, the A-Lamp network produces higher score for the original image than both transformed versions. Figure \ref{trans_img} shows examples used in the study and their transformed versions, along with the A-Lamp predicted posteriors. The result shows that the A-Lamp network is able to reliably respond to the change of image layout and fine-grained details caused by the transformations. 
	In addition, we also notice that when the image content is more semantic, it will be more sensitive to holistic layout. In particular, the warped version of the portrait photo receives much lower score than the original one, or even the down-sampled one. It is interesting to notice that the warped version for the second photo example seems not so bad, while the down-sampled version falls a lot due to much detail loss. To further investigate the effectiveness of our A-Lamp networks adaption for content-based image aesthetics, we have performed content-based photo aesthetic study with detailed results presented in the next.

	\subsection{Content-based photo aesthetic analysis} \label{content}
	To carry out content-based photo aesthetic study, we take photos in eight most popular semantic tags used in \cite{Murray:MMP2012:AVA}: portrait, animal, still-life, food-drink, architecture, floral, cityscape and landscape. We used the same testing image collection used in \cite{Lu:2014:TMM:rating}, approximately 2.5K for testing in each of the categories. In each of the eight categories, we systematically compared New-MP-Net and A-Lamp network with the baseline approach \cite{Murray:MMP2012:AVA} (denoted by AVA) and the state-of-the-art approach in \cite{Lu:2014:TMM:rating}. Specifically, SCNN{\tiny c} and SCNN{\tiny w} denote the single-column CNN in \cite{Lu:2014:TMM:rating} that takes center cropping and warping, respectively, as inputs. DCNN denotes the double-column CNN in \cite{Lu:2014:TMM:rating}. As shown in Figure.\ref{category}, the proposed network training approach significantly outperforms the state-of-the-art in most of the categories, where ”floral” and ”architecture” show substantial improvements. We find that, photos belonging to these two categories often show complicated texture details, which can be seen in Figure \ref{samples}. The proposed adaptive Multi-Patch subnet keeps the fine-grained details and thus produces much better performance. We also find that A-Lamp networks shows much better performance than New-MP-Net in ”portrait” and ”animal”. These results indicate that once an image is associated with a clear semantic meaning, then the global view is more important than the local views in terms of assessing image aesthetics. Figure\ref{samples} shows some examples of the test images that are considered by the proposed A-Lamp as among the highest and lowest aesthetics values. These photos are selected from all eight categories.
	
	\section{Conclusion}
	This paper presents an Adaptive Layout-Aware Multi-Patch Convolutional Neural Network (A-Lamp CNN) architecture for photo aesthetic assessment. This novel scheme is able to accept arbitrary sized images and to capture intrinsic aesthetic characteristics from both fined grained details and holistic image layout simultaneously. To support A-Lamp training on these hybrid inputs, we  developed a dedicated double-subnet neural network structure, i.e. a Multi-Patch subnet and a Layout-Aware subnet. We then construct an aggregation layer to effectively combine the hybrid features from these two subnets. Extensive experiments on the large-scale AVA benchmark show that this A-Lamp CNN can significantly improve the state of the art in photo aesthetics assessment. Meanwhile, the proposed A-Lamp CNN can be directly applied to many other computer vision tasks, such as style classification, object category recognition, image retrieval, and scene classification, which we leave as our future work.
	
	\cleardoublepage
	{\small
		\bibliographystyle{IEEE}
		\bibliography{test}
	}
	
\end{document}